\documentclass{article}

\usepackage{PRIMEarxiv}
\usepackage{amsmath}
\usepackage[utf8]{inputenc} % allow utf-8 input
\usepackage[T1]{fontenc}    % use 8-bit T1 fonts
\usepackage{hyperref}       % hyperlinks
\usepackage{url}            % simple URL typesetting
\usepackage{booktabs}       % professional-quality tables
\usepackage{amsfonts}       % blackboard math symbols
\usepackage{nicefrac}       % compact symbols for 1/2, etc.
\usepackage{microtype}      % microtypography
\usepackage{lipsum}
\usepackage{fancyhdr}       % header
\usepackage{graphicx}       % graphics
\graphicspath{{media/}}     % organize your images and other figures under media/ folder

\title{Classification and Morphological Analysis of DLBCL Subtypes in H\&E-Stained Slides
}

\author{
Ravi Kant Gupta$\star$ \hspace{5mm} Mohit Jindal$\star$ \hspace{5mm} Garima Jain$\dagger$  \hspace{5mm} Epari Sridhar$\diamond$\hspace{5mm} \textbf{Subhash Yadav}$\diamond$ \\ \textbf{Hasmukh Jain}$\diamond$\hspace{5mm}
\textbf{Tanuja Shet}$\diamond$\hspace{5mm}
\textbf{Uma Sakhdeo}$\diamond$\hspace{5mm}
\textbf{Manju Sengar}$\diamond$\hspace{5mm}
\textbf{Lingaraj Nayak}$\diamond$\\
\textbf{Bhausaheb Bagal}$\diamond$\hspace{5mm}
\textbf{Umesh Apkare}$\diamond$\hspace{5mm}
\textbf{Amit Sethi}$\star$\\
  $\star$Department of Electrical Engineering,
  Indian Institute of Technology Bombay \\
  $\dagger$Koita Center for Digital Health, Indian Institute of Technology Bombay \\
  $\diamond$Tata Memorial hospital, HBNI
  Mumbai, India\\
  \texttt{\{ravigupta131, 21D070068, asethi\}@iitb.ac.in} \\
}

\begin{document}
\maketitle

\begin{abstract}
We address the challenge of automated classification of diffuse large B-cell lymphoma (DLBCL) into its two primary subtypes: activated B-cell-like (ABC) and germinal center B-cell-like (GCB). Accurate classification between these subtypes is essential for determining the appropriate therapeutic strategy, given their distinct molecular profiles and treatment responses. Our proposed deep learning model demonstrates robust performance, achieving an average area under the curve (AUC) of (87.4 ± 5.7)\% during cross-validation. It shows a high positive predictive value (PPV), highlighting its potential for clinical application, such as triaging for molecular testing. To gain biological insights, we performed an analysis of morphological features of ABC and GCB subtypes. We segmented cell nuclei using a pre-trained deep neural network and compared the statistics of geometric and color features for ABC and GCB. We found that the distributions of these features were not very different for the two subtypes, which suggests that the visual differences between them are more subtle. These results underscore the potential of our method to assist in more precise subtype classification and can contribute to improved treatment management and outcomes for patients of DLBCL.
\end{abstract}

% keywords can be removed
\keywords{classification, deep learning, lymphoma, morphological, subtype.}

\section{Introduction}
Diffuse large B-cell lymphoma (DLBCL) comprises a collection of aggressive B-cell non-Hodgkin lymphomas (NHL), characterized by significant genetic diversity and a wide range of clinical manifestations~\cite{roschewski2020molecular}. It is the most common subtype, accounting for 30-40\% of all B-cell NHL cases. The median age at diagnosis is  seventh decade and patients present with widespread lymph-adenopathy ~\cite{vodicka2022diffuse}.
More than 20 years ago, gene expression profiling (GEP) identified two distinct molecular subtypes, classified according to cell of origin (COO), the germinal centre B-cell-like (GCB) subtype and the non-GCB subtype, which includes activated B-cell-like (ABC) and primary mediastinal B-cell subtypes. The COO classification categorizes subtypes that originate from B-cells at various stages of development, each characterized by unique oncogenic mechanisms, dependence on distinct survival pathways, and varying clinical outcomes~\cite{roschewski2020molecular}. These DLBCL subtypes exhibit significant differences in prognosis, with ABC subtype linked to poorer outcomes~\cite{nowakowski2015abc}.

DLBCL is a potentially curable disease. Standard first-line immunochemotherapy, which  typically includes rituximab along with cyclophosphamide, doxorubicin, vincristine, and prednisone (R-CHOP), produces remission rates of 60-70\%. It was seen that patients with ABC subtype treated with the R-CHOP regimen tend to have worse outcomes compared to those with the GCB subtype, with a 5-year progression-free survival (PFS) of 48\% versus 73\% . With the advent of novel therapies i.e. immunomodulatory agents, like ibrutinib or lenalidomide, The response rate in ABC subtype has improved to 95\%~\cite{nowakowski2015abc}.

GEP by cDNA microarray is the gold standard for identifying the molecular subtypes of DLBCL. However, due to ease of availability and cost, traditionally, immunohistochemistry is used for classification. The Hans algorithm, uses antibodies for CD10, BCL6, and IRF4/MUM1. However the identified subgroups do not perfectly align with molecular categories~\cite{nguyen2020recommendations}.

\begin{figure} 
\centering
\includegraphics[height=5.5cm,width=8.5cm]{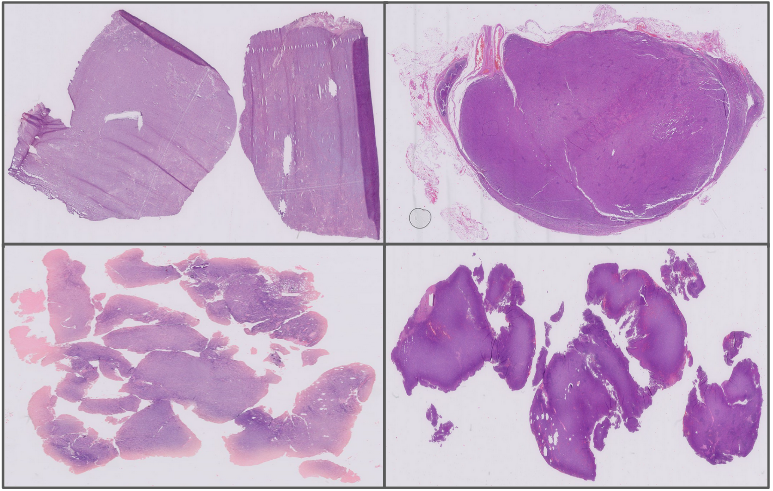}
\caption{Snapshot of  H\&E stained images from TMC Dataset: Top row are images of GCB subtype and bottom row, ABC subtypes of DLBCL respectively}
\label{sample_thumbnail}
\end{figure}

\begin{figure*} 
\centering
\includegraphics[height=6cm,width=15cm]{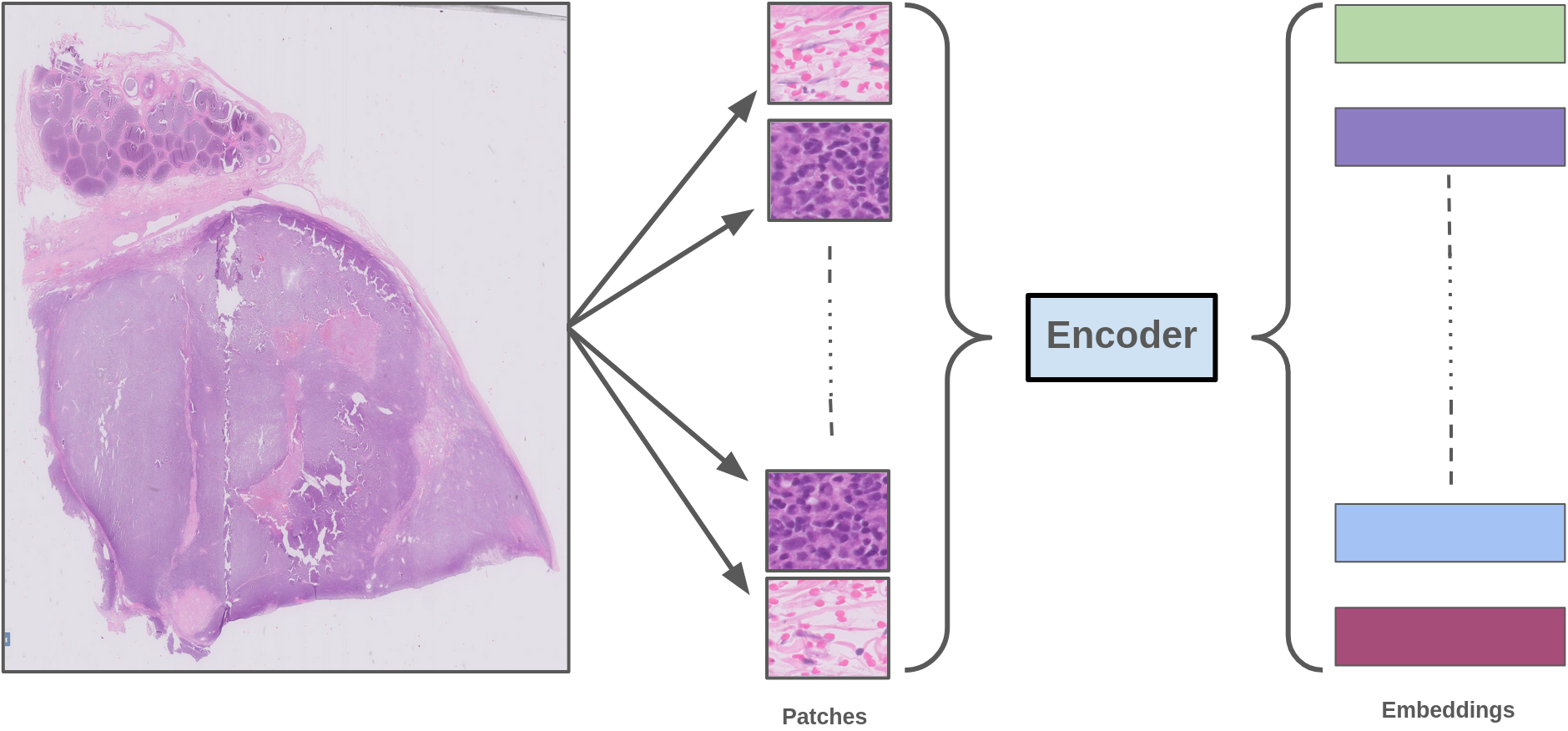}
\caption{Patches are generated from H\&E stained WSI with optimal preprocessing and fed to encoder (pretrained) to get the embedding.}
\label{preprocessing}
\end{figure*}

\begin{figure*} 
\centering
\includegraphics[height=5cm,width=16cm]{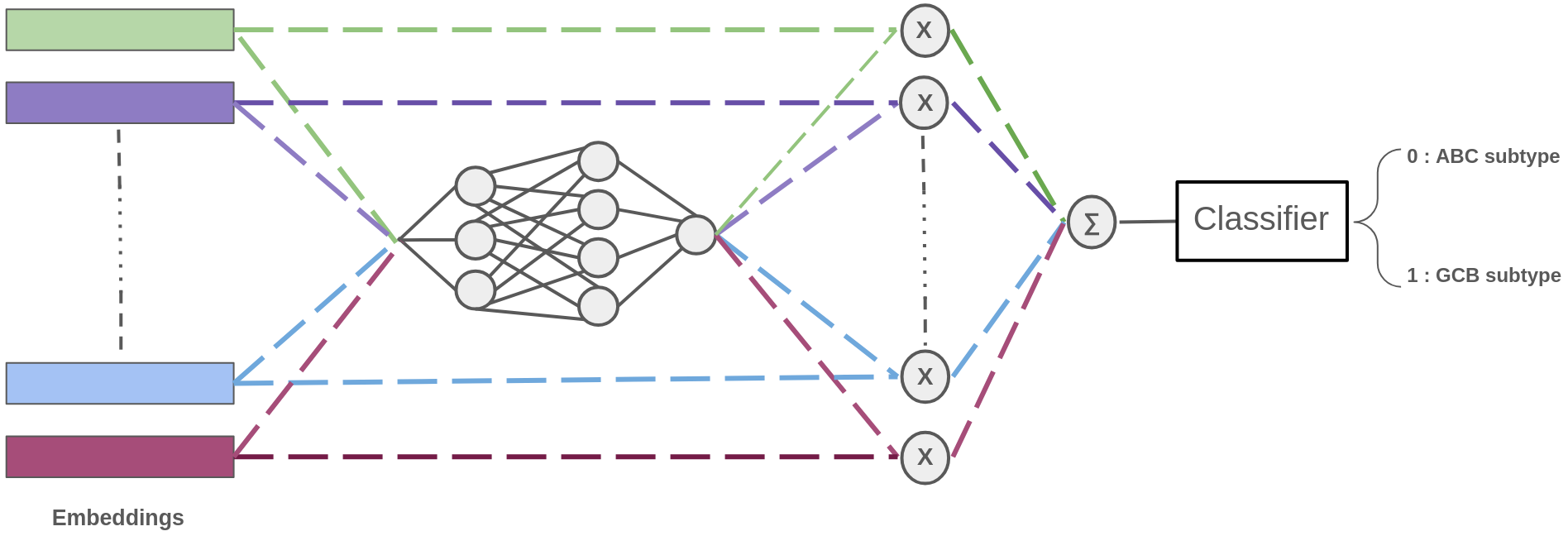}
\caption{Schematic of the attention-based classification model. The features are processed through an attention block (NN), where importance weights are applied to each feature. The weighted features are then aggregated and passed to a classifier.}
\label{workflow}
\end{figure*}

To tackle this classification challenge between activated B-cell-like (ABC) and germinal center B-cell-like (GCB) subtypes, we implemented a weakly supervised technique such as the customized version~\cite{gupta2023egfr} of CLAM~\cite{lu2021data} (Clustering-constrained Attention Multiple-instance learning) and MIL~\cite{ilse2018attention}(Multiple Instance Learning) model, which is particularly effective for analyzing whole slide images. We experimented with a variety of feature extractors, including ResNet34~\cite{jian2016deep}, ResNet50~\cite{jian2016deep}, RegNet~\cite{7995968}, ConvNeXT\_Tiny~\cite{liu2022convnet}, EfficientNet~\cite{tan2019efficientnet}, and Swin\_Tiny~\cite{liu2021swin}, each chosen for their ability to capture different levels of histological detail. In conjunction with these feature extractors, we applied a range of histology-specific augmentations designed to simulate realistic variations in the images, such as alterations in staining intensity, rotation, and scaling. These augmentations helped to enhance the robustness and generalizability of the model. By leveraging the CLAM model alongside these advanced feature extractors and augmentation strategies, we aimed to significantly improve the accuracy and reliability of subtype classification.

To substantiate our claims and rigorously evaluate the proposed methodology, we employed the TMC Dataset, a proprietary collection comprising 115 whole slide images (WSIs). Each image in this dataset has been meticulously annotated and classified, providing a robust foundation for validating our approach. The TMC Dataset offers a diverse range of histological patterns, representative of the challenges in accurately classifying subtypes such as activated B-cell-like (ABC) and germinal center B-cell-like (GCB) lymphomas. To provide a visual context, Fig. 1 displays the thumbnails of the WSIs included in the dataset, showcasing the complex and varied histopathological features that our model must process and classify. The use of these thumbnails illustrates the dataset's complexity, emphasizing the technical rigor required in our model’s analysis and the critical importance of accurate subtype differentiation in this challenging domain.

\section{Related Work}
Given the clinical implications, several classification algorithms have been developed, some utilizing widely accessible immunohistochemical methods. While these methods show high concordance with gene expression profiling (GEP) and hold significant prognostic value, the underlying technical and biological assumptions are yet to be optimized to enhance classification accuracy and performance. ~\cite{hans2004confirmation} introduced the first algorithm based on three protein markers: CD10, BCL6, and MUM1. This algorithm classified patients into two groups (GCB and Non-GCB)  but has low concordance with GEP analysis (71\% for GCB and 88\% for Non-GCB)  and yielded inconsistent results for patients treated with R-CHOP in terms of prognostic significance. Another algorithm, proposed by ~\cite{choi2009new}, incorporated two additional markers, FOXP1 and GCET1, but also showed limited concordance (83\%) with GEP analysis when differentiating between GCB and Non-GCB subtypes. ~\cite{visco2012comprehensive} developed a more effective method known as the Visco-Young algorithm, which demonstrated high concordance (92.6\%) between patients with GCB and ABC gene profiles. This algorithm, which included MME, FOXP1, and BCL6, exhibited strong independent prognostic value in DLBCL patients treated with R-CHOP. 

Despite its growing use in clinical practice, some limitations hinder its widespread adoption. The immunohistochemical staining process is sensitive to numerous factors, including the skill of the experimenter, which introduces subjectivity in interpreting results.
In recent years, machine learning is being increasingly used for COO subclassification of DLBCL. ~\cite{zhao2016machine} developed a new algorithm for subclassification of DLBCL using gene expression profiles from five GEO databases, validating its accuracy. The model, based on the differential expression of eight genes (e.g., MYBL1, IRF4), showed high sensitivity and specificity in distinguishing between GCB and ABC subtypes. Another algorithm was developed to classify subtypes of DLBCL with an accuracy of 91.6\% when compared to gene expression profiling (GEP). The algorithm demonstrated a true positive rate of 95.7\% for GCB and 87\% for ABC, with a high area under the ROC curve (0.934).  The Perfecto-Villela (PV) algorithm~\cite{perfecto2019discriminant} leverages machine learning techniques to classify DLBCL into GCB and non-GCB subtypes using IHC data. The PV algorithm was developed by applying Bayesian classifiers, support vector machines (SVM), and artificial neural networks (ANN) to a dataset containing immunohistochemical staining information. By focusing on key antibodies (CD10, FOXP1, GCET1, and MUM1), the PV algorithm achieved 94\% accuracy, 93\% specificity, and 95\% sensitivity, outperforming traditional algorithms like Hans and Choi. ~\cite{carreras2021artificial} demonstrated that artificial neural networks (ANNs) can accurately predict molecular subtypes of DLBCL using a pancancer immune-oncology gene panel. The multilayer perceptron (MLP) model achieved an area under the curve (AUC) of 0.98 for OS prediction and correctly classified the molecular subtypes with an accuracy of 99\% in the training dataset and 95\% in the testing dataset. 

Prediction of DLBCL subtype on histopathology images using machine learning is being increasingly worked on and has shown promising results. ~\cite{yuce2023pb2377} developed deep-learning model to classify cell-of-origin (COO) using digitized H\&E -stained slides. The tile-based model achieved an AUC of 0.73 and 0.67 for the test and holdout sets, respectively, while the nucleus-based model had AUCs of 0.63 and 0.70. The study found that the model identified significant morphological differences, such as higher tumour cell density and larger cell size in ABC compared to GCB DLBCL, which could serve as novel features for COO classification. Deep learning model  developed by ~\cite{vrabac2021dlbcl} identified morphological features from DLBCL histology sections that correlate with patient survival outcomes. The model achieved a C-index of 0.635 (95\% CI: 0.574-0.691) using only geometric features derived from tumor nuclei, which indicates better-than-random survival prediction. When combined with clinical features, the model's C-index improved to 0.700 (95\% CI: 0.651-0.744), suggesting that integrating geometric and clinical features enhances prognostic accuracy.

\section{Methodology}

In the preprocessing stage of our methodology, we addressed the significant challenges posed by handling large, gigapixel-sized whole slide images (WSIs) and the associated artifacts. To mitigate these issues, we applied a deep learning model~\cite{PATIL2023100306} to effectively remove artifacts. Once we identified the usable tissue regions within the WSIs, we divided them into manageable patches of 256x256 pixels at a 40x zoom level. We extracted 256x256 pixel patches shown in Fig. 2, balancing the need to preserve histological detail with computational efficiency while minimizing overlap to ensure a diverse representation of the tissue features. This process involved several critical steps: first, tissue detection was performed to focus on the relevant regions of interest; next, white patches, which represent non-informative areas, were filtered out. To ensure that only patches with meaningful content were analyzed, we excluded patches with low cellularity, specifically those containing 0 to 10 nuclei using HistomicsTK library~\cite{HistomicsTK}. These carefully selected patches were then fed into a variety of backbone encoders, including ResNet34~\cite{jian2016deep}, ResNet50~\cite{jian2016deep}, RegNet~\cite{7995968}, ConvNeXT\_Tiny~\cite{liu2022convnet}, EfficientNet~\cite{tan2019efficientnet}, and Swin\_Tiny~\cite{liu2021swin}, to extract robust and informative feature embeddings (of size \(\in \mathbb{R}^{1024}\)) for classification task.

After extracting the feature embeddings \(\mathbf{e}_k \in \mathbb{R}^{1024}\) for each patch \(i\) from a whole slide image (WSI) using various backbone encoders, these embeddings are processed through the first fully-connected layer $\mathbf{W}_1 \in \mathbb{R}^{512 \times 1024}$ further compresses each fixed 1024-dimensional patch-level representation $\mathbf{e}_k$ to a 512-dimensional vector $\mathbf{h}_k = \mathbf{W}_1 \mathbf{e}_k^\top$.
Then this is passed through an attention mechanism to enhance the classification of the entire slide. We consider the first two layers of the attention network $\mathbf{U}_a \in \mathbb{R}^{256 \times 512}$ and $\mathbf{V}_a \in \mathbb{R}^{256 \times 512}$ collectively as the attention backbone shared by all classes. The attention network then splits into two parallel attention branches $\mathbf{W}_{a,1}, \mathbf{W}_{a,2} \in \mathbb{R}^{1 \times 256}$.
The attention mechanism assigns an attention score \(a_{k,m}\) to each patch(here kth instance from one wsi and mth class), indicating its relative importance in the context of the slide-level classification task.
The attention scores \(a_{k,m}\) are computed as follows:
% \[
% \alpha_i = \frac{\exp(\mathbf{w}^\top \tanh(\mathbf{V}\mathbf{E}_i + \mathbf{b}))}{\sum_{j=1}^{N} \exp(\mathbf{w}^\top \tanh(\mathbf{V}\mathbf{E}_j + \mathbf{b}))}
% \]

\begin{equation}
a_{k,m} = \frac{\exp\left\{ \mathbf{W}_{a,m} \left( \tanh \left( \mathbf{V}_a \mathbf{h}_k^\top \right) \odot \text{sigm} \left( \mathbf{U}_a \mathbf{h}_k^\top \right) \right) \right\}}{\sum_{j=1}^N \exp\left\{ \mathbf{W}_{a,m} \left( \tanh \left( \mathbf{V}_a \mathbf{h}_j^\top \right) \odot \text{sigm} \left( \mathbf{U}_a \mathbf{h}_j^\top \right) \right) \right\}}
\end{equation}

This attention-based aggregation enables the model to focus on the most informative patches. It mitigates the impact of irrelevant or noisy regions, thereby improving the accuracy and robustness of the subtype classification.
The slide-level representation aggregated per the attention score distribution for the mth class is, 

% The weighted sum of the patch embeddings, denoted by \(\mathbf{z}\), is then computed using the attention scores:

\begin{equation}
\mathbf{h}_{\text{slide},m} = \sum_{k=1}^N a_{k,m} \mathbf{h}_k
\end{equation}

% \(h_{slide,m}\) is then passed through a classifier layer $\mathbf{W}_{c,m} \in \mathbb{R}^{1 \times 512}$ to get unnormalised slide level score \(h_{slide,m}\), and it is shown as,
% $s_{\text{slide},m} = \mathbf{W}_{c,m} \mathbf{h}_{\text{slide},m}^\top$.
% To get the prediction probabilities to slide-level, we applied softmax function on top of \(s_{slide}\) 

% \[
% \mathbf{z} = \sum_{i=1}^{N} \alpha_i \mathbf{E}_i
% \]

This aggregated feature vector \(h_{slide,m}\) is used as the input for the final classifier, which determines the subtype classification of the entire slide. Specifically, the classifier predicts ($ s_{slide} $) whether the slide belongs to the ABC or GCB subtype:

\begin{equation}
s_{\text{slide}} = \sigma(\mathbf{W}_c \mathbf{h}_{\text{slide},m}^\top + \mathbf{b}_c)
\end{equation}

where \(\mathbf{W}_c\) and \(b_c\) are the weights and bias of the classifier, respectively, and \(\sigma\) represents the softmax function that outputs the probability distribution over the classes.

 Fig. 3 depicts our approach that effectively synthesizes localized information across patches into a cohesive global representation, leading to a more precise classification of the WSI as belonging to either the ABC or GCB subtype.

\section{Dataset and Implementation Details}
\subsection{Dataset}

\begin{figure*} 
\centering
\includegraphics[height=12cm,width=15cm]{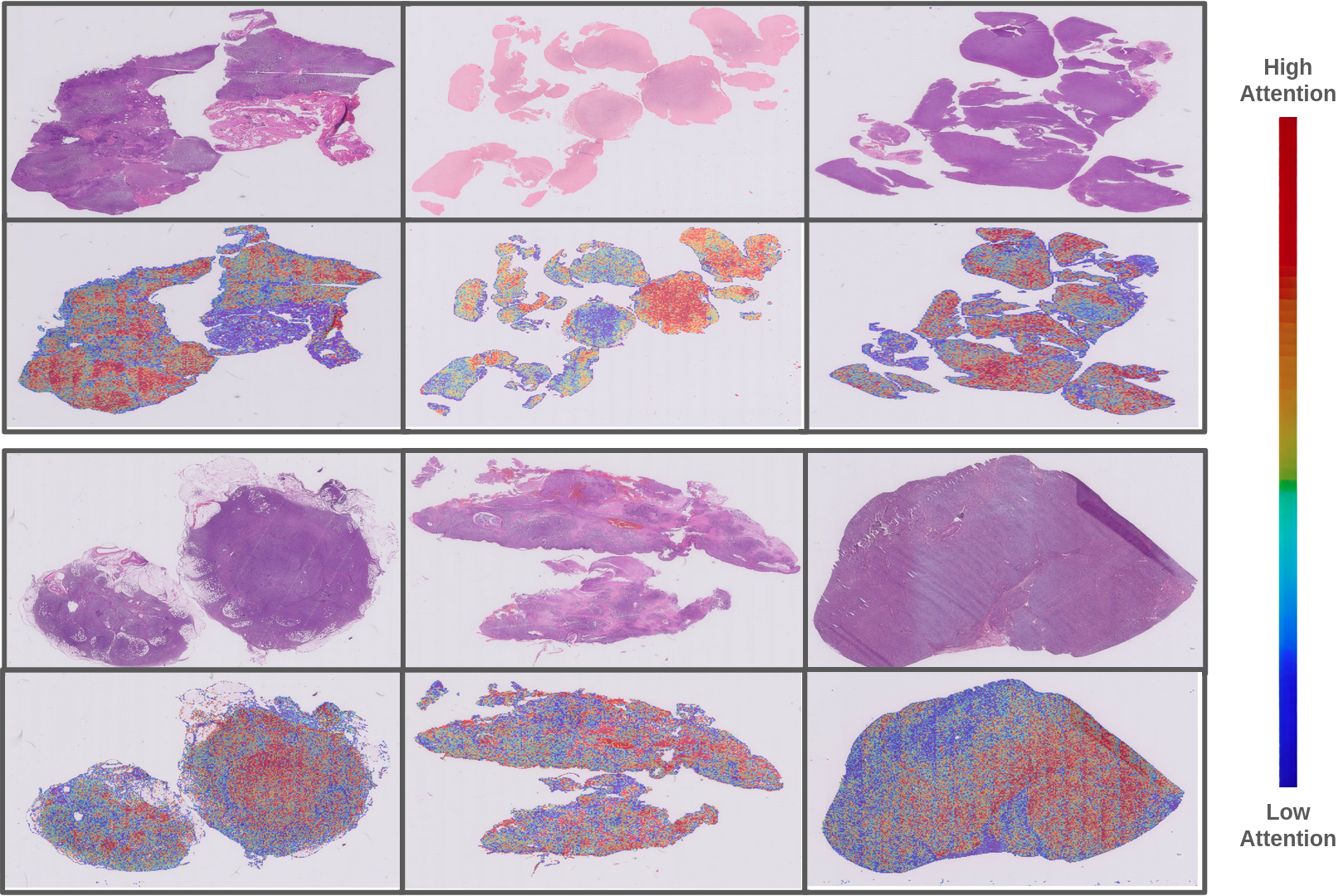}
\caption{The first row shows thumbnails of three WSIs from the ABC subtype, with corresponding heatmaps in the second row (red indicates high attention, blue indicates low attention). The third and fourth rows display the same for the GCB subtype.}
\label{heatmap}
\end{figure*}

To conduct our study we have used an Indian dataset from the Tata Memorial Centre (TMC), Mumbai. The work on TMC was approved by the TMC Institutional Ethics Commmittee. This dataset consist of H\&E stained large (gigapixel size) WSIs of lymph-node tissue. The dataset included 115 whole slide images from 115 different patients. All slides are labeled as ABC subtype or GCB subtype of DLBCL. 
% The NOS label was used when the classification of an image was ambiguous and could belong to either the ABC or GCB classes. Due to this ambiguity, images labeled as NOS were excluded from further analysis to ensure clarity in the dataset.\\
% After excluding the NOS-labeled images, the updated dataset consisted of 115 images. 
The dataset consists of 115 whole slide images derived from lymph node biopsies, with 62 images classified under the activated B-cell-like (ABC) subtype and the remaining 53 classified under the germinal center B-cell-like (GCB) subtype. To rigorously assess the performance of our model, we employed a three-fold cross-validation strategy. In this approach, the dataset was systematically partitioned into three distinct folds, each containing a separate split of training, validation, and testing subsets. For each fold, 70\% of the images were used for training, 15\% for validation, and 15\% for testing. This careful division ensures that each subtype is adequately represented across all subsets, allowing for robust evaluation and minimizing the risk of over fitting or bias in the results.

\begin{figure*}
\centering
\includegraphics[height=10cm,width=15cm]{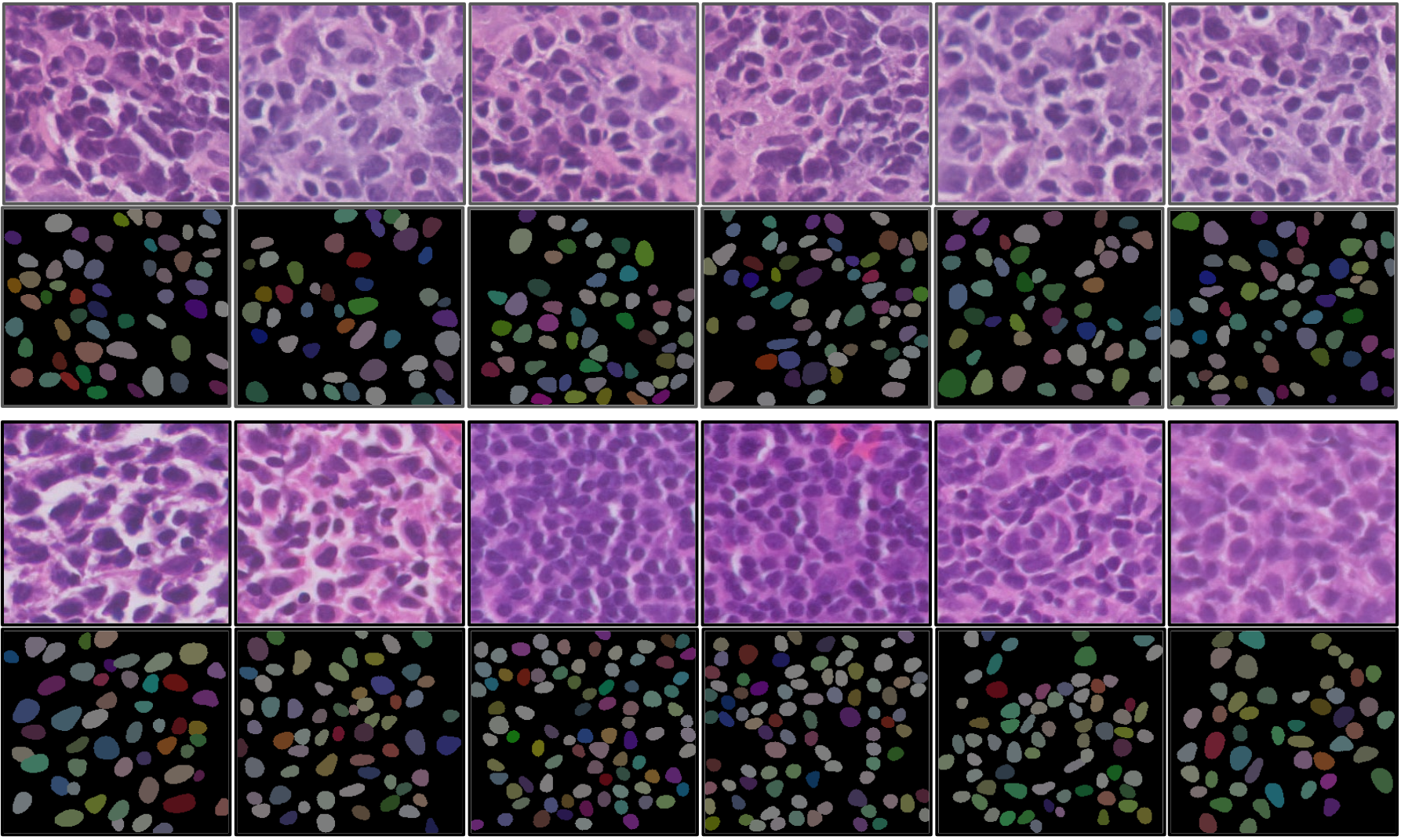}
\caption{The top row displays the highly attended patches of the ABC class from multiple patients, followed by the second row, which shows their corresponding segmentation maps obtained from StarDist. The third row presents the highly attended patches of the GCB class from multiple patients, with the last row depicting the respective segmentation maps for these GCB patches}
\label{seg}
\end{figure*}

\subsection{Implementation Details}
All the experiments were conducted on an NVIDIA A100 in PyTorch, using multiple backbones pre-trained on ImageNet~\cite{deng2009imagenet} as the backbone for our proposed model. Multiple experiments were conducted with different hyperparameters to achieve the best results. The final weight decay is 0.00004 with a learning rate of 0.0002 and a dropout of 0.1. We train models by a maximum of 200 epochs, however, we use early stopping criteria of 20 epochs to avoid unnecessary time in training. We have used AdamW~\cite{loshchilov2019decoupled} with a momentum of 0.9, and a weight decay of 0.00004 as the optimizer. We adhere to the standard procedure for unsupervised classification, wherein we use labeled training samples and unlabeled test samples during the training process.  For a fair comparison with prior works, we also conduct experiments with the same backbones for demonstration of results with different datasets.

% \begin{table}
% \caption{Performance comparison on the DLBL dataset (3 fold) with different backbones as feature extractor and CLAM model as a classifier}
% \begin{center}
% \begin{tabular}{|l|c|c|c|c|}
% \hline
% Parameter  & P value & ABC & GCB & F test\\
% \hline
% Nuclear Area   & 1.623e-20 & Low & High & p vale - 1.801e-124\\
% \hline
% Nuclear Perimeter &  1.502e-20 & Low & High\\
% \hline
% Aspect Ratio  & 0.313 & High & Low\\
% \hline
% Cell Count  & 9.869e-231 & High & Low\\
% \hline
% Bounding Box Area  & 1.026e-25 & Low & High \\
% \hline
% Circularity & 1.276e-14 & High & Low \\
% \hline
% Extent  & 3.127e-10 & High & Low \\
% \hline
% Solidity  & 8.320e-9 & Same & Same\\
% \hline
% NC Ratio & 1.020e-64 & Low & High\\
% \hline
% RB Ratio & 2.064e-29 & Low & High\\
% \hline
% \end{tabular}
% \end{center}  
% \label{table2}
% \end{table}

\begin{figure*}
\centering
\includegraphics[height=9cm,width=15cm]{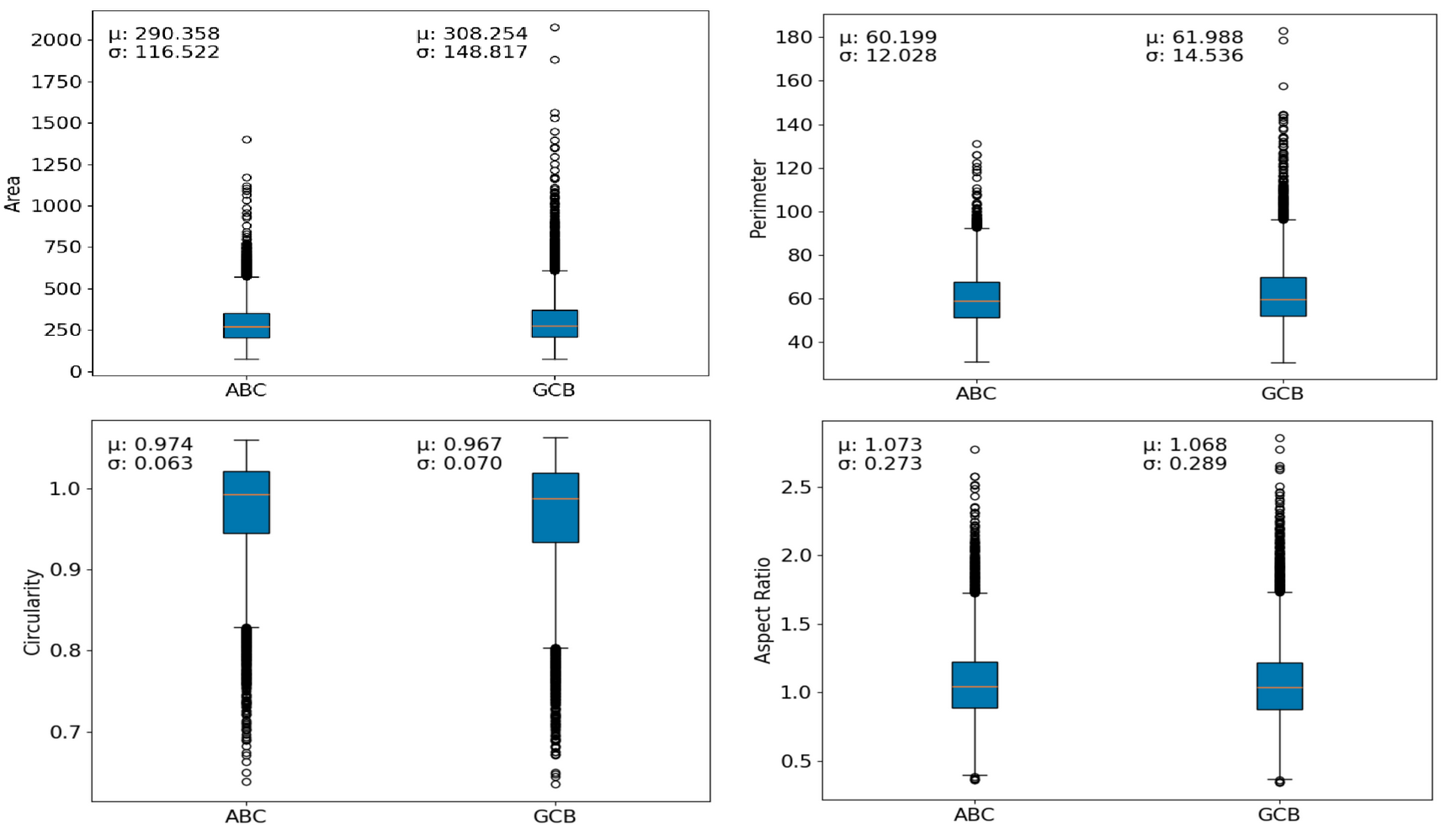}
\caption{The plots show that the ABC subtype has a slightly lower mean ($\mu$) nucleus area and slightly higher aspect ratio, while the GCB subtype exhibits slightly higher variability in shape, as indicated by the standard deviation ($\sigma$) in circularity and aspect ratio. The mean and standard deviation are written in each plot, providing insight into the morphological differences between the two subtypes.}
\label{box1}
\end{figure*}

\begin{figure*} 
\centering
\includegraphics[height=9cm,width=15cm]{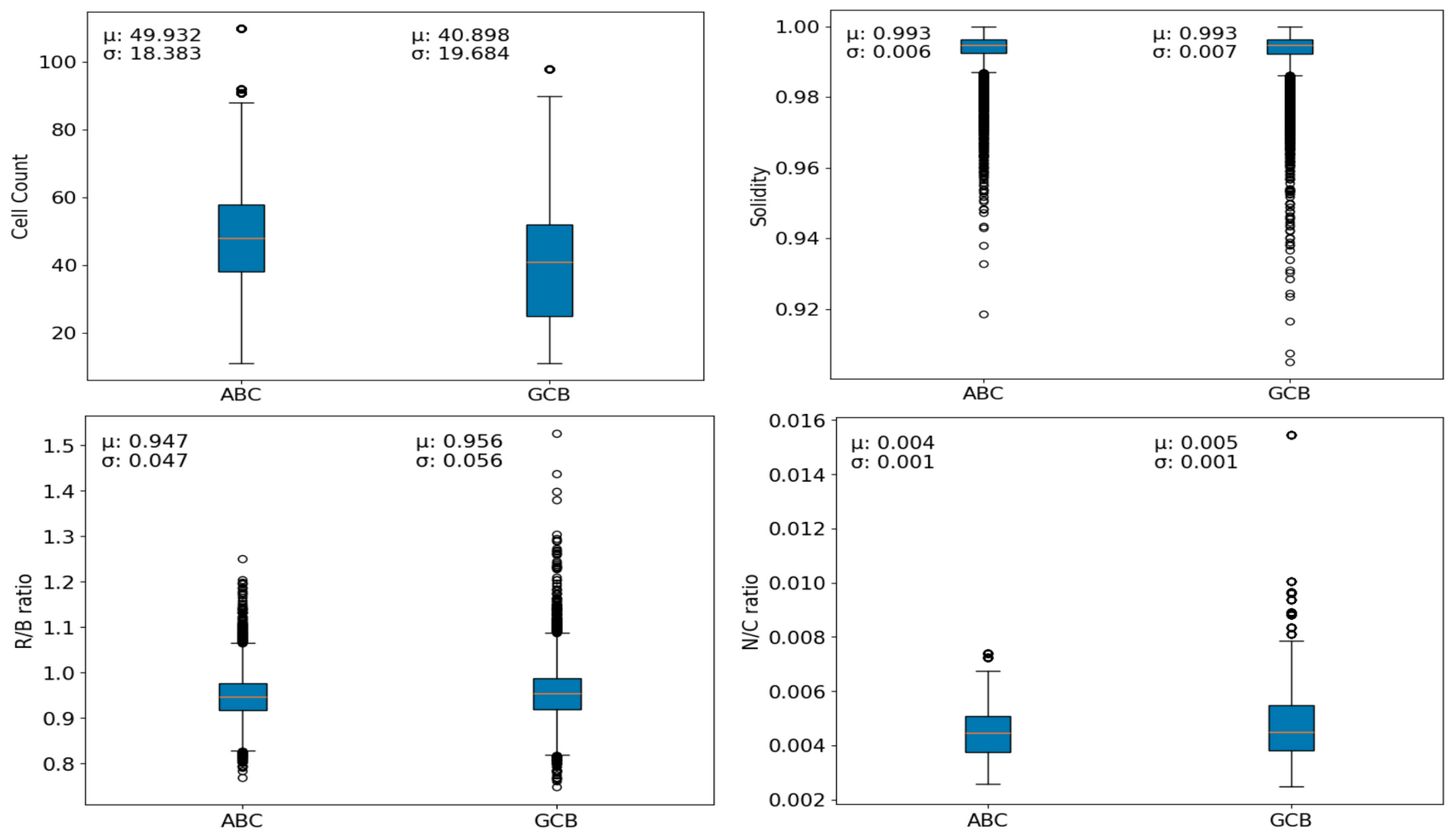}
\caption{The plots show that the ABC subtype has a significantly higher mean ($\mu$) cell count and slightly lower R/B ratio, while the GCB subtype exhibits slightly higher variability in general, as indicated by the standard deviation ($\sigma$) in all the 4 plots. The mean and standard deviation are written in each plot, providing insight into the morphological differences between the two subtypes.}
\label{elbow}
\end{figure*}

\section{Results and Discussion}

\begin{table*}[ht!]
\caption{Performance comparison on the TMC dataset (3 fold) with different pretrained backbones as feature extractor and CLAM model as a classifier}
\begin{center}
\resizebox{0.95\textwidth}{!}{% Adjusting to 95% of the text width
\begin{tabular}{|l|c|c|c|c|c|}
\hline
Backbone Model	 &	Avg. AUC$\pm$ STD \% &	Max AUC \% & Avg ACC $\pm$ STD \% & Max ACC \% \\
\hline
ResNet34~\cite{jian2016deep}  & 77.2 $\pm$ 8.0 & 84.2 & 67.6 $\pm$ 6.0 & 76.4\\
\hline
ResNet50~\cite{jian2016deep}  & 77.6 $\pm$ 2.0 & 81.0& 61 $\pm$ 2.9 &  65.0\\
\hline
RegNet~\cite{7995968}  & 79.8 $\pm$ 1.3 & 81.4 & 65.7 $\pm$ 5.5 & 73.3\\
\hline
EfficientNet~\cite{tan2019efficientnet}  & 72.0 $\pm$ 6.4 & 78.5 & 57.3 $\pm$ 4.7 & 64.0\\
\hline
Swin\_Tiny~\cite{liu2021swin} & 74.2 $\pm$ 12.2 &	84.2 & 63.3 $\pm$ 4.7 & 70.5 \\
\hline
ConvNeXT\_Tiny~\cite{liu2022convnet} & \textbf{87.4 $\pm$ 5.7} & \textbf{94.4} & \textbf{79.3 $\pm$ 5.1} & \textbf{86.6} \\
\hline
\end{tabular}
}
\end{center}  
\label{table2}
\end{table*}

Our study employed an attention-based Multiple Instance Learning (MIL) framework to classify ABC and GCB lymphoma subtypes, achieving an impressive average AUC of 87.4 $\pm$ 5.7 \% across three-fold cross-validation shows in Table~\ref{table2}. In exploring various backbone encoders, we found that the ConvNeXT\_Tiny encoder consistently provided the most robust features, significantly enhancing the model's performance. The attention mechanism within the MIL framework demonstrated its effectiveness by consistently focusing on tumor regions, even when the entire tissue was exposed to the model. The heatmaps generated for the high attention patches, as identified by our model, are presented in Fig. 4. These heatmaps illustrate the regions of interest for both ABC and GCB classes, highlighting the areas(patches of 256x256 pixel size) within the tissue region of WSI that were most critical for the model's classification decisions. For WSIs classified as ABC, the heatmaps show areas of high attention in red and low attention in blue, with the color scale ranging from blue (low attention) to red (high attention). A similar color coding is applied to the heatmaps for WSIs classified as GCB. This focus on the most relevant areas of the whole slide images (WSIs) is a crucial aspect of our approach, ensuring that the model directs its learning capacity toward the most informative regions for subtype classification.

% \begin{table*}[ht!]
% \caption{Performance comparison on the TMC dataset (3 fold) with different pretrained backbones as feature extractor and CLAM model as a classifier}
% \begin{center}
% \resizebox{\textwidth}{!}{%
% \begin{tabular}{|l|c|c|c|c|c|}
% \hline
% Backbone Model	 &	Avg. AUC$\pm$ STD &	Max AUC  & Avg ACC	$\pm$ STD& Max ACC\\
% \hline
% ResNet34~\cite{}  & 77.2 $\pm$ 8 & \textbf{84.2} & 67.6 $\pm$ 6 & 76.4\\
% \hline
% ResNet50~\cite{}  & 77.6 $\pm$ 2 & \textbf{81}& 61 $\pm$ 2.9 &  65\\
% \hline
% RegNet~\cite{}  & 79.8 $\pm$ 1.3 & \textbf{81.4} & 65.7 $\pm$ 5.5 & 73.3\\
% \hline
% EfficientNet~\cite{}  & 72 $\pm$ 6.4 & \textbf{78.5} & 57.3 $\pm$ 4.7 & 64\\
% \hline
% Swin\_Tiny~\cite{} & 74.2 $\pm$ 12.2 &	\textbf{84.2} & 63.3 $\pm$ 4.7 & 70.5 \\
% \hline
% ConvNext\_Tiny~\cite{} & 87.4 $\pm$ 5.7 & \textbf{94.4} & 79.3 $\pm$ 5.1 & 86.6 \\
% \hline
% \end{tabular}
% }
% \end{center}  
% \label{table2}
% \end{table*}

The highly attended patches were processed using the StarDist~\cite{schmidt2018cell} model to obtain segmentation of each nucleus, as illustrated in Fig. 5 (the top row shows six highly attended patches from the ABC class with corresponding segmentation maps in the second row, and similarly for the GCB class in the third and fourth rows). The segmentation results confirmed that the attention mechanism had effectively selected patches with significant cellular content, further supporting the reliability of our morphological analysis. These combined findings highlight the strength of our approach in distinguishing between ABC and GCB lymphoma subtypes. Further analysis of these segmented patches revealed several important morphological differences between the ABC and GCB subtypes. To systematically quantify these differences, we conducted a detailed analysis using box plots, where each data point represented an individual cell from the respective subtype. Initially, our dataset consisted of 115 WSIs, but we selectively used 54 slides based on the results of StarDist segmentation inference, ensuring the highest quality and relevance for our analysis. Of these, 23 WSIs were from the ABC subtype and 31 from the GCB subtype. From each WSI, we extracted 10 representative patches with highest attention, resulting in a comprehensive and robust dataset with high number of cells. Notably, the ABC subtype exhibited a slightly lower mean nucleus area and perimeter compared to the GCB subtype, indicating that the cells in ABC patches might be more compact. However, ABC patches showed a higher mean nucleus count, suggesting that these patches are more densely populated with cells, a potentially significant feature in distinguishing between these subtypes. To validate these observations, we performed p-value calculations for each morphological feature. For example, the p-value for nucleus area was \(1.623 \times 10^{-20}\), for perimeter \(1.502 \times 10^{-20}\), and for cell count \(9.869 \times 10^{-231}\), indicating that these differences are statistically significant. In contrast, the aspect ratio had a p-value of 0.313, suggesting that the observed difference in aspect ratio between the subtypes may not be statistically significant. The use of p-value calculations allows us to rigorously validate our hypothesis by statistically confirming whether the observed differences in morphological features between ABC and GCB subtypes are likely due to chance or represent true biological distinctions. While the circularity values were almost similar between the subtypes, the GCB subtype displayed slightly more variability in shape, as indicated by a higher standard deviation in circularity, aspect ratio and other parameters. Solidity was uniformly high across both subtypes, reflecting that the cells in these patches are generally compact and well-defined. These findings are visually represented in the box plot shown in Fig. 6.

% The p-values for bounding box area \(1.026 \times 10^{-25}\), circularity \(1.276 \times 10^{-14}\), extent \(3.127 \times 10^{-10}\), solidity \(8.320 \times 10^{-9}\), and N/C ratio \(1.020 \times 10^{-64}\) further confirm the significance of the differences in these features.

The R/B ratio, which measures the relative intensity of the red and blue color channels, and the N/C ratio, indicating the nuclear-to-cytoplasmic area ratio, were nearly identical between the subtypes (R/B ratio p-value: \(2.064 \times 10^{-29}\), N/C ratio p-value: \(1.020 \times 10^{-64}\)). Thus Fig. 7 suggests that the differences in staining intensity and nuclear-cytoplasmic proportions are minimal and may not be significant distinguishing features. However, the consistent focus of the attention mechanism on tumor regions, as demonstrated by the heatmaps, underscores the model's ability to isolate and emphasize the most relevant areas for classification.

The integration of attention mechanisms with the ConvNeXT\_Tiny encoder has proven to be particularly effective, offering a nuanced understanding of the morphological characteristics that differentiate these subtypes, and potentially improving diagnostic accuracy in clinical applications.

% \section{Conclusion}
\section{Conclusion}
In conclusion, this study demonstrates the effectiveness of an attention-based Multiple Instance Learning (MIL) framework combined with advanced feature extraction techniques for the classification of DLBCL subtypes, specifically distinguishing between germinal center B-cell-like (GCB) and activated B-cell-like (ABC) subtypes. By integrating backbone encoders like ConvNeXT\_Tiny and applying histology-specific augmentations, the proposed model achieved notable accuracy and robustness, as evidenced by an average AUC of 87.4 ± 5.7\% in three-fold cross-validation. The study’s results are further validated through heatmaps and segmentation analyses, which reveal key morphological differences between the subtypes, such as variations in nucleus area, aspect ratio, and cell count, that are statistically significant.

 This approach offers a scalable, interpretable solution for digital pathology in DLBCL, providing an alternative to traditional gene expression profiling (GEP) and immunohistochemistry (IHC) methods that can be limited by accessibility and consistency. In a test set of 20 images, consisting of 10 ABC and 10 GCB cases, the model achieved a positive predictive value (PPV) of 90.0\% and a negative predictive value (NPV) of 70.0\%, further demonstrating its effectiveness in distinguishing between these lymphoma subtypes. The attention mechanism effectively highlights tumor-rich regions, enhancing the model’s focus on the most relevant histological features. Our model identifies morpho-molecular features that can act as substitutes for GEP to predict the DLBCL subtype. This has widespread applications in resource-poor countries, where molecular tests are not equitably available, accessible, and affordable. As digital pathology continues to evolve, integrating such advanced machine learning models holds significant potential for improving diagnostic accuracy and guiding personalized treatment strategies in DLBCL, ultimately contributing to better patient outcomes.

%Bibliography
\bibliographystyle{unsrt}
\bibliography{references}

\begin{thebibliography}{10}

\bibitem{roschewski2020molecular}
Mark Roschewski, James~D Phelan, and Wyndham~H Wilson.
\newblock Molecular classification and treatment of diffuse large b-cell lymphoma and primary mediastinal b-cell lymphoma.
\newblock {\em The Cancer Journal}, 26(3):195--205, 2020.

\bibitem{vodicka2022diffuse}
Prokop Vodicka, Pavel Klener, and Marek Trneny.
\newblock Diffuse large b-cell lymphoma (dlbcl): early patient management and emerging treatment options.
\newblock {\em OncoTargets and Therapy}, 15:1481, 2022.

\bibitem{nowakowski2015abc}
Grzegorz~S Nowakowski and Myron~S Czuczman.
\newblock Abc, gcb, and double-hit diffuse large b-cell lymphoma: does subtype make a difference in therapy selection?
\newblock {\em American Society of Clinical Oncology Educational Book}, 35(1):e449--e457, 2015.

\bibitem{nguyen2020recommendations}
Florence Nguyen-Khac, Audrey Bidet, Lauren Veronese, Agnes Daudignon, Dominique Penther, Marie-B{\'e}reng{\`e}re Troadec, Christine Lefebvre, and Marina Lafage-Pochitaloff.
\newblock Recommendations for cytogenomic analysis of hematologic malignancies: comments from the francophone group of hematological cytogenetics (gfch).
\newblock {\em Leukemia}, 34(6):1711--1713, 2020.

\bibitem{gupta2023egfr}
Ravi~Kant Gupta, Shivani Nandgaonkar, Nikhil~Cherian Kurian, Tripti Bameta, Subhash Yadav, Rajiv~Kumar Kaushal, Swapnil Rane, and Amit Sethi.
\newblock Egfr mutation prediction of lung biopsy images using deep learning.
\newblock 2023.

\bibitem{lu2021data}
Ming~Y Lu, Drew~FK Williamson, Tiffany~Y Chen, Richard~J Chen, Matteo Barbieri, and Faisal Mahmood.
\newblock Data-efficient and weakly supervised computational pathology on whole-slide images.
\newblock {\em Nature biomedical engineering}, 5(6):555--570, 2021.

\bibitem{ilse2018attention}
Maximilian Ilse, Jakub Tomczak, and Max Welling.
\newblock Attention-based deep multiple instance learning.
\newblock In {\em International conference on machine learning}, pages 2127--2136. PMLR, 2018.

\bibitem{jian2016deep}
S~Jian, H~Kaiming, R~Shaoqing, and Z~Xiangyu.
\newblock Deep residual learning for image recognition.
\newblock In {\em IEEE Conference on Computer Vision \& Pattern Recognition}, pages 770--778, 2016.

\bibitem{7995968}
Nick Schneider, Florian Piewak, Christoph Stiller, and Uwe Franke.
\newblock Regnet: Multimodal sensor registration using deep neural networks.
\newblock In {\em 2017 IEEE Intelligent Vehicles Symposium (IV)}, pages 1803--1810, 2017.

\bibitem{liu2022convnet}
Zhuang Liu, Hanzi Mao, Chao-Yuan Wu, Christoph Feichtenhofer, Trevor Darrell, and Saining Xie.
\newblock A convnet for the 2020s.
\newblock In {\em Proceedings of the IEEE/CVF conference on computer vision and pattern recognition}, pages 11976--11986, 2022.

\bibitem{tan2019efficientnet}
Mingxing Tan and Quoc Le.
\newblock Efficientnet: Rethinking model scaling for convolutional neural networks.
\newblock In {\em International conference on machine learning}, pages 6105--6114. PMLR, 2019.

\bibitem{liu2021swin}
Ze~Liu, Yutong Lin, Yue Cao, Han Hu, Yixuan Wei, Zheng Zhang, Stephen Lin, and Baining Guo.
\newblock Swin transformer: Hierarchical vision transformer using shifted windows.
\newblock In {\em Proceedings of the IEEE/CVF international conference on computer vision}, pages 10012--10022, 2021.

\bibitem{hans2004confirmation}
Christine~P Hans, Dennis~D Weisenburger, Timothy~C Greiner, Randy~D Gascoyne, Jan Delabie, German Ott, H~Konrad Muller-Hermelink, Elias Campo, Rita~M Braziel, Elaine~S Jaffe, et~al.
\newblock Confirmation of the molecular classification of diffuse large b-cell lymphoma by immunohistochemistry using a tissue microarray.
\newblock {\em Blood}, 103(1):275--282, 2004.

\bibitem{choi2009new}
William~WL Choi, Dennis~D Weisenburger, Timothy~C Greiner, Miguel~A Piris, Alison~H Banham, Jan Delabie, Rita~M Braziel, Huimin Geng, Javeed Iqbal, Georg Lenz, et~al.
\newblock A new immunostain algorithm classifies diffuse large b-cell lymphoma into molecular subtypes with high accuracy.
\newblock {\em Clinical cancer research}, 15(17):5494--5502, 2009.

\bibitem{visco2012comprehensive}
Carlo Visco, Yan Li, Zijun~Y Xu-Monette, Roberto~N Miranda, Tina~M Green, A~Tzankov, W~Wen, WM~Liu, BS~Kahl, ESG d'Amore, et~al.
\newblock Comprehensive gene expression profiling and immunohistochemical studies support application of immunophenotypic algorithm for molecular subtype classification in diffuse large b-cell lymphoma: a report from the international dlbcl rituximab-chop consortium program study.
\newblock {\em Leukemia}, 26(9):2103--2113, 2012.

\bibitem{zhao2016machine}
Shuangtao Zhao, Xiaoli Dong, Wenzhi Shen, Zhen Ye, and Rong Xiang.
\newblock Machine learning-based classification of diffuse large b-cell lymphoma patients by eight gene expression profiles.
\newblock {\em Cancer medicine}, 5(5):837--852, 2016.

\bibitem{perfecto2019discriminant}
Yocanx{\'o}chitl Perfecto-Avalos, Alejandro Garcia-Gonzalez, Ana Hernandez-Reynoso, Gildardo S{\'a}nchez-Ante, Carlos Ortiz-Hidalgo, Sean-Patrick Scott, Rita~Q Fuentes-Aguilar, Ricardo Diaz-Dominguez, Grettel Le{\'o}n-Mart{\'\i}nez, Ver{\'o}nica Velasco-Vales, et~al.
\newblock Discriminant analysis and machine learning approach for evaluating and improving the performance of immunohistochemical algorithms for coo classification of dlbcl.
\newblock {\em Journal of translational medicine}, 17:1--12, 2019.

\bibitem{carreras2021artificial}
Joaquim Carreras, Shinichiro Hiraiwa, Yara~Yukie Kikuti, Masashi Miyaoka, Sakura Tomita, Haruka Ikoma, Atsushi Ito, Yusuke Kondo, Giovanna Roncador, Juan~F Garcia, et~al.
\newblock Artificial neural networks predicted the overall survival and molecular subtypes of diffuse large b-cell lymphoma using a pancancer immune-oncology panel.
\newblock {\em Cancers}, 13(24):6384, 2021.

\bibitem{yuce2023pb2377}
Anil Yuce, Jacob Gildenblat, Samaneh Abbasi-Sureshjani, Camille Laurent, and Konstanty Korski.
\newblock Pb2377: Predicting cell of origin from digitized images of hematoxylin and eosin-stained slides of diffuse large b-cell lymphomas using a cell-based deep-learning model.
\newblock {\em HemaSphere}, 7(S3):e5900850, 2023.

\bibitem{vrabac2021dlbcl}
Damir Vrabac, Akshay Smit, Rebecca Rojansky, Yasodha Natkunam, Ranjana~H Advani, Andrew~Y Ng, Sebastian Fernandez-Pol, and Pranav Rajpurkar.
\newblock Dlbcl-morph: morphological features computed using deep learning for an annotated digital dlbcl image set.
\newblock {\em Scientific Data}, 8(1):135, 2021.

\bibitem{PATIL2023100306}
Abhijeet Patil, Harsh Diwakar, Jay Sawant, Nikhil~Cherian Kurian, Subhash Yadav, Swapnil Rane, Tripti Bameta, and Amit Sethi.
\newblock Efficient quality control of whole slide pathology images with human-in-the-loop training.
\newblock {\em Journal of Pathology Informatics}, 14:100306, 2023.

\bibitem{HistomicsTK}
Histomicstk.
\newblock \url{https://github.com/DigitalSlideArchive/HistomicsTK}.
\newblock Accessed: [12/10/2023].

\bibitem{deng2009imagenet}
Jia Deng, Wei Dong, Richard Socher, Li-Jia Li, Kai Li, and Li~Fei-Fei.
\newblock Imagenet: A large-scale hierarchical image database.
\newblock In {\em 2009 IEEE conference on computer vision and pattern recognition}, pages 248--255. Ieee, 2009.

\bibitem{loshchilov2019decoupled}
Ilya Loshchilov and Frank Hutter.
\newblock Decoupled weight decay regularization.
\newblock {\em arXiv preprint arXiv:1711.05101}, 2017.

\bibitem{schmidt2018cell}
Uwe Schmidt, Martin Weigert, Cameron Broaddus, and Gene Myers.
\newblock Cell detection with star-convex polygons.
\newblock {\em International Conference on Medical Image Computing and Computer-Assisted Intervention}, pages 265--273, 2018.

\end{thebibliography}

\end{document}